\documentclass[journal]{IEEEtran}
\usepackage{latexsym}

\usepackage{enumerate}
\usepackage{graphicx}
\usepackage{subfigure}
\usepackage{multirow}

\usepackage{framed,multirow}
\usepackage{times}
\usepackage{graphicx} 
\usepackage{subfigure}
\usepackage{amssymb}
\usepackage{latexsym}

\usepackage{algorithm}
\usepackage{algorithmic}

\usepackage{hyperref}

\usepackage{url}
\usepackage{xcolor}
\usepackage{amsmath}
\usepackage{amsxtra}
\usepackage{braket}
\usepackage{array}
\usepackage{color}
\usepackage{url}
\usepackage{placeins}

\usepackage{times}
\usepackage{balance}
\usepackage{enumerate}
\usepackage{multirow}
\usepackage{pifont}
\usepackage{marvosym}
\usepackage{ifsym}
\usepackage{threeparttable}

\usepackage{lipsum} 
\usepackage{amsmath,esint} 
\usepackage{cuted}

\begin{document}
\title{AERK: Aligned Entropic Reproducing Kernels through Continuous-time Quantum Walks}
\author{Lixin~Cui,~\IEEEmembership{}
        Ming~Li,~\IEEEmembership{}
        Yue~Wang,~\IEEEmembership{}
        Lu~Bai${}^{*}$,~\IEEEmembership{}
        Edwin R.~Hancock,~\IEEEmembership{IEEE~Fellow}

\thanks{Lixin Cui and Yue Wang are with Central University of Finance and Economics, Beijing, China. Ming Li is with Department of Computer Science, Zhejiang Normal University, Zhejiang, China. Lu Bai (${}^{*}$Corresponding Author: bailu@bnu.edu.cn; bailucs@cufe.edu.cn) is with School of Artificial Intelligence, Beijing Normal University, Beijing, China, and Central University of Finance and Economics, Beijing, China. Edwin R. Hancock is with Department
of Computer Science, University of York, York, UK}
}
\markboth{IEEE Transactions on ...}%
{Shell \MakeLowercase{\textit{et al.}}: Bare Demo of IEEEtran.cls for Journals}

\maketitle

\begin{abstract}
In this work, we develop an Aligned Entropic Reproducing Kernel (AERK) for graph classification. We commence by performing the Continuous-time Quantum Walk (CTQW) on each graph structure, and computing the Averaged Mixing Matrix (AMM) to describe how the CTQW visit all vertices from a starting vertex. More specifically, we show how this AMM matrix allows us to compute a quantum Shannon entropy for each vertex of a graph. For pairwise graphs, the proposed AERK kernel is defined by computing a reproducing kernel based similarity between the quantum Shannon entropies of their each pair of aligned vertices. The analysis of theoretical properties reveals that the proposed AERK kernel cannot only address the shortcoming of neglecting the structural correspondence information between graphs arising in most existing R-convolution graph kernels, but also overcome the problem of neglecting the structural differences between pairs of aligned vertices arising in existing vertex-based matching kernels. Moreover, unlike existing classical graph kernels that only focus on the global or local structural information of graphs, the proposed AERK kernel can simultaneously capture both global and local structural information through the quantum Shannon entropies, reflecting more precise kernel based similarity measures between pairs of graphs. The above theoretical properties explain the effectiveness of the proposed kernel. The experimental evaluation on standard graph datasets demonstrates that the proposed AERK kernel is able to outperform state-of-the-art graph kernels for graph classification tasks.
\end{abstract}


\maketitle
\IEEEpeerreviewmaketitle

\section{Introduction}\label{s1}

In recent years, graph-based representations are extensively employed in the region of pattern recognition and machine learning for analyzing structure data, e.g., the social networks~\cite{DBLP:conf/pkdd/Bai0BH19}, image-based graphs~\cite{DBLP:journals/jmiv/BaiH13}, 3D shape-based graphs~\cite{DBLP:conf/cvpr/EscolanoHL11}, molecule networks~\cite{DBLP:conf/nips/GasteigerBG21}, traffic networks~\cite{DBLP:journals/apin/ZengPHYH22}, etc. With this context, there has been
incremental interests in utilizing graph kernel methods associated with a specific kernel machine
(e.g. the C-Support Vector Machine (C-SVM), etc) for applications of graph classification~\cite{GarterCOLT2003,KonderProductKernel}.

\subsection{Literature Review}\label{s1.1}

A graph kernel is usually defined in terms of a similarity measure between graph structures. By now, most state-of-the-art graph kernels essentially fall into the concept of R-convolution, which was first defined by Haussler~\cite{haussler99convolution} in 1999. This is a generic manner of defining a novel graph kernel based on measuring the similarity between the decomposed substructures of graphs. Specifically, for each pair of
graphs $G_p(V_p,E_p)$ and $G_q(V_q,E_q)$ where $V_{\{\cdot\}}$ corresponds to the vertex set and $E_{\{\cdot\}}$ corresponds to the edge set, let $\mathcal{S}_{p}\sqsubseteq G_p$ and $\mathcal{S}_{q}\sqsubseteq G_q$ be any pair of substructures based on a specific graph decomposing approach, a R-convolution graph kernel $k_R$ between
the pairwise graphs $G_p$ and $G_q$ is defined as
\begin{align}
k_R(G_p,G_q)=\sum_{\mathcal{S}_{p}\sqsubseteq G_p}  \sum_{\mathcal{S}_{q}\sqsubseteq G_q}  s(\mathcal{S}_{p},\mathcal{S}_{q}), \nonumber
\end{align}
where $s(\mathcal{S}_{p},\mathcal{S}_{q})$ is a similarity measure between the substructures
$\mathcal{S}_{p}$ and $\mathcal{S}_{q}$. Generally, $s$ is usually set as the Dirac kernel, and $s(\mathcal{S}_{p},\mathcal{S}_{q})$ equals to $1$ if $\mathcal{S}_{p}$ and $\mathcal{S}_{q}$ are isomorphic to each other (i.e., $\mathcal{S}_{p} \simeq \mathcal{S}_{q}$ ), and $0$ otherwise.

With the concept of R-convolution, one can employ any possible graph decomposing approach to define a new R-convolution based graph kernel, e.g., the kernels based on decomposed (a) paths~\cite{DBLP:conf/icdm/BorgwardtK05}, (b) cycles~\cite{DBLP:journals/tnn/AzizWH13}, (c) walks~\cite{DBLP:conf/nips/SugiyamaB15}, (d) subgraphs~\cite{DBLP:conf/icml/KriegeM12}, (e) subtrees~\cite{DBLP:journals/jmlr/AzaisI20}, etc. For instance, Kalofolias et al.,~\cite{DBLP:conf/sdm/KalofoliasWV21} have introduced a Structural Similarity Random Walk Kernel based on the similarity measure between random walk based substructures. Borgwardt et al.,~\cite{DBLP:conf/icdm/BorgwardtK05} have introduced a Shortest-Path Kernel by calculating the number of pairwise shortest paths with different same lengths, that are extracted by the classical Floyd Method~\cite{DBLP:journals/cacm/Floyd62a}. Aziz et al.,~\cite{DBLP:journals/tnn/AzizWH13} have introduced a non-Backtrackless Kernel by counting the pairs of the cycles with different same lengths, that are identified by the Ihara zeta function~\cite{DBLP:journals/tnn/RenWH11}. Costa and Grave~\cite{DBLP:conf/icml/CostaG10} have introduced a Pairwise-Distance Expansion Subgraph Kernel based on sets of layer-wise expansion subgraphs around pairwise rooted vertices, that have different appointed distances between each other. Shervashidze et al.,~\cite{shervashidze2010weisfeiler} have introduced a Weisfeiler-Lehman Subtree Kernel based on the subtree-based substructures, that are extracted by the classical Weisfeiler-Lehman Isomorphism Test Method~\cite{UWL}. Other effective R-convolution graph kernels also include: (1) the Pyramid-Quantized Shortest-Path Kernel~\cite{DBLP:journals/ijon/GkirtzouB16}, (2) the Pyramid-Quantized Weisfeiler-Lehman Subtree Kernel~\cite{DBLP:journals/ijon/GkirtzouB16}, (3) the Wasserstein Weisfeiler-Lehman Subtree Kernel~\cite{DBLP:conf/nips/TogninalliGLRB19}, (4) Marginalized Random-Walk Kernel~\cite{DBLP:conf/icml/KashimaTI03}, (5) the Isolation Graph Kernel~\cite{DBLP:conf/aaai/XuTJ21}, (6) the Graph Filtration Kernel~\cite{DBLP:conf/aaai/SchulzWW22}, (7) the Subgraph Alignment Kernel~\cite{DBLP:conf/icml/KriegeM12}, (8) the Binary-Subtree Kernel~\cite{DBLP:conf/bmvc/GaidonHS11}, etc.

\begin{figure}
\centering
\subfigure{\includegraphics[width=1.0\linewidth]{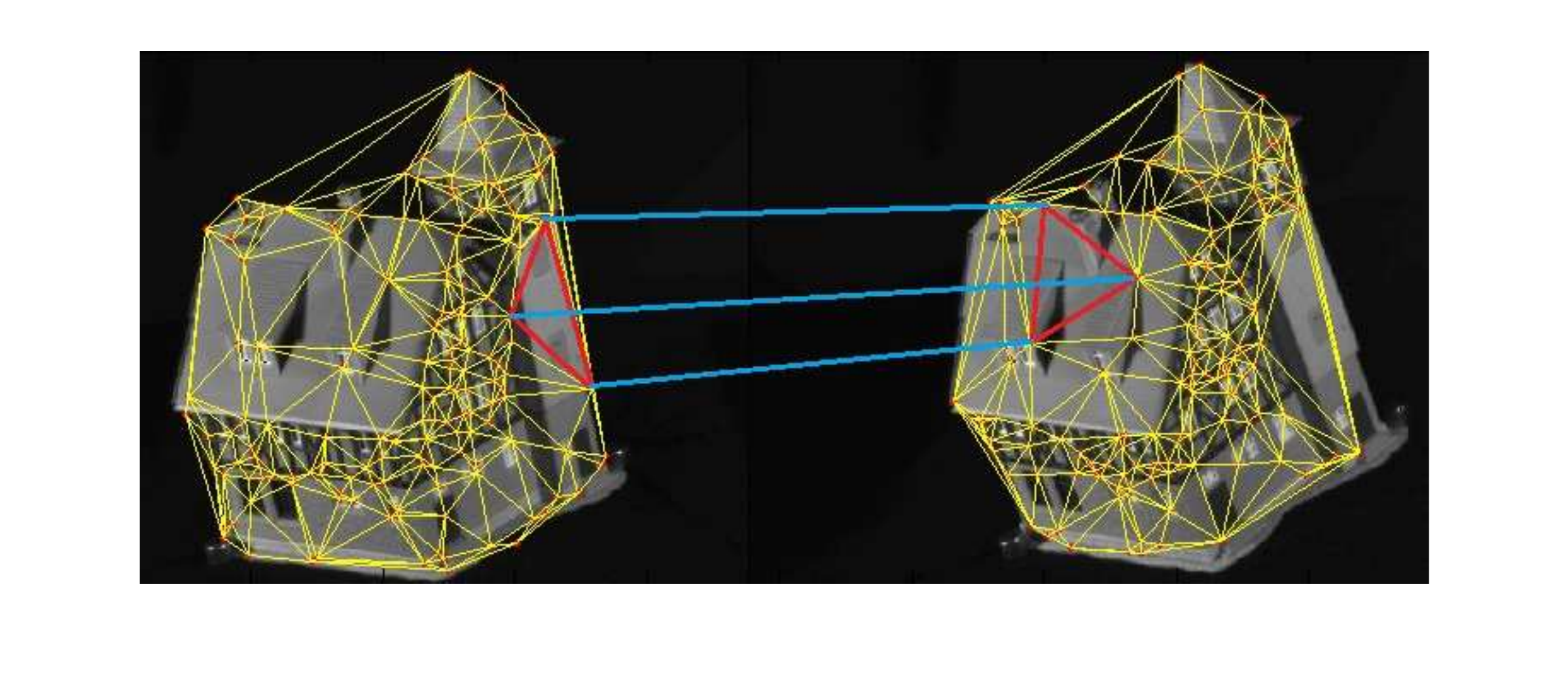}}
\vspace{-30pt}
\caption{\footnotesize{Delaunay graphs extracted from the same house object through different viewpoints.}} \label{cvpr}
\vspace{-20pt}
\end{figure}

Although, the above classical R-convolution graph kernels have demonstrated their performance on graph classification tasks, associated with standard real-world datasets. Most of them still suffer form two common inherent drawbacks, due to the R-convolution framework. First, measuring the similarity or isomorphism between the substructures of large sizes is always a challenging problem, and usually requires expensive computational complexity. To guarantee the computational efficiency, the R-convolution graph kernels tend to utilize small sized substructures, that only reflect local structural information. As a result, the R-convolution graph kernels fail to capture characteristics of the global graph structures. Second, the R-convolution graph kernels only focus on whether a pair of substructures are isomorphic, and each pair of isomorphic substructures will contribute one unit kernel value. In other words, the R-convolution graph kernels completely ignore the structural correspondence information between the substructures, in terms of the global graph structures. For example, Fig.~\ref{cvpr} exhibits two Delaunay graphs, that are both extracted from the same house object through different viewpoints. Since the two triangle-based substructures are isomorphic, the R-convolution graph kernels will directly add an unit kernel value, no matter whether the substructures are structurally aligned to each other in terms of the vision background (i.e., the global graph structures). As a result, the R-convolution graph kernels can not reflect precise similarity measures between graphs. Overall, the above two problems significantly affect the effectiveness of the R-convolution graph kernels.

To address the drawback of ignoring global structure information that arising in existing R-convolution graph kernels, a number of global-based graph kernels that focus on capture global characteristics of graphs through the global graph structure have been developed. Under this scenario, Johansson et al.,~\cite{DBLP:conf/icml/JohanssonJDB14} have introduced a Global Geometric Embedding Kernel associated with the Lov\'asz numbers as well as their orthonormal representations, that are computed through the adjacency matrix. Bai and Hancock et al.,~\cite{DBLP:journals/jmiv/BaiH13} have introduced a Jensen-Shannon Graph Kernel based on measuring the classical Jensen-Shannon Divergence (JSD) between both the approximated von Neumann entropies and the random walk based Shannon entropies of global graph structures. Xu et al.,~\cite{DBLP:journals/pr/XuJBXL18} have introduced a Global-based Reproducing Graph Kernel by measuring the Basic Reproducing Kernel between the approximated von Neumann entropies of global graph structures. On the other side, to further capture more complicated global structure information, Bai and Rossi et al.,~\cite{DBLP:journals/pr/Bai0TH15,rossi2015measuring} have developed a family of quantum-inspired global-based graph kernels, namely the Quantum Jensen-Shannon Graph Kernels, by measuring the Quantum Jensen-Shannon Divergence (QJSD) between the density operators of the Continuous-Time Quantum Walk (CTQW)~\cite{DBLP:journals/pr/EmmsWH09} evolved on global graph structures. Different from the classical Continuous-Time Random Walk (CTRW)~\cite{DBLP:conf/sigmetrics/RibeiroFST11} that is dominated by the doubly stochastic matrix, the CTQW is controlled by the unitary matrix and not leaded with the low Laplacian spectrum frequencies. Thus, the CTQW can not only better discriminate different graphs, but also reflect intrinsic structural characteristics. As a result, the resulting Quantum Jesen-Shannon Graph Kernels can capture more complicated intrinsic structure information residing on global graphs. Unfortunately, the above global-based graph kernels focus more on the global characteristics, and ignore the inherent characteristics residing on local structures. Furthermore, similar to the R-convolution graph kernels, these global-based graph kernels cannot identify the structural correspondence information between pairwise graphs, influencing the effective kernel-based similarity measures.

To resolve the drawback of overlooking the correspondence information that arises in both the R-convolution graph kernels and the global-based graph kernels, Bai and Xu et al.,~\cite{DBLP:conf/icml/Bai0ZH15,DBLP:conf/ijcai/BaiZW0H15,DBLP:journals/pr/XuBJTZL21} have developed a family of vertex-based matching kernels by counting the pairs of aligned vertices, that are identified by evaluating the distance between the vectorial depth-based representations~\cite{DBLP:journals/pr/BaiH14} of vertices in a Euclidean space. Since each pair of aligned vertices theoretically correspond to a pair of approximately isomorphic subgraphs. These matching kernels are theoretically equivalent to aligned subgraph kernels that calculate the number of pairwise structurally-aligned subgraphs, encapsulating the structural correspondence information between graphs. However, similar to the R-convolution graph kernels, these matching kernels cannot capture global structure information through the local aligned vertices, influencing the effectiveness of the kernel measures. Furthermore, since each pair of aligned vertices will dedicate the same one unit kernel value, the matching kernels also cannot identify structural differences between different pairs of aligned vertices in terms of global graph structures, influencing the precise kernel measure. Finally, these matching kernels only identify the correspondence information based on the specific dimensional vectorial representations of vertices, lacking multilevel alignment information between graphs.

Overall, the above literature review in terms of the R-convolution graph kernels, global-based graph kernels, and the vertex-based matching kernels reveals that developing effective graph kernels is always a theoretical challenging problem.

\subsection{Contributions}\label{s1.2}
The main objective of this work is to resolve the theoretical problems of the existing graph kernel methods discussed above. To this end, we develop a novel Aligned Entropic Reproducing Kernel (AERK) based on the Averaged Mixing Matrix (AMM) of the CTQW, for graph classification applications. One of the key innovations for the proposed AERK kernels is to compute the kernel value by measuring the similarity between the quantum Shannon entropies of aligned vertices, through the AMM matrix of the CTQW. Since these entropies of the vertices are different, the proposed AERK kernel can discriminate the structural difference between different pairs of aligned vertices based on the entropic similarity. Overall, the contributions of this work are summarized as follows.

\textbf{First}, for each pair of graphs, we commence by performing the CTQW on each of the graph structures. More specifically, we utilize the AMM matrix to describe the evolution of the CTQW. The reason of using the AMM matrix is due to the fact that it can not only reflect richer graph topological information in terms of the CTQW, but also assign each vertex an individual probability distribution that investigates how the CTQW visit all vertices when it departs from the vertex (see details in Section~\ref{s3.1}). Furthermore, we show how the AMM matrix allows us to compute a quantum Shannon entropy for each vertex, that not only reflects local structural information residing on the local vertex but also capture complicated intrinsic structural characteristics of global graph structures.

\textbf{Second}, for the pair of graphs we construct a family of different $h$-level entropic correspondence matrices to reflect multilevel alignment information between the vertices. With the entropic correspondence matrices as well as the quantum Shannon entropies of vertices to hand, the proposed AERK kernel is defined by computing the sum of the reproducing kernel-based similarities between the entropies over all pairs of aligned vertices. We theoretically demonstrate that the proposed AERK kernel cannot only address the shortcoming of ignoring structural correspondence information between graphs arising in existing R-convolution graph kernels, but also overcome the problem of neglecting the structural differences between pairs of aligned vertices arising in existing vertex-based matching kernels. Moreover, unlike the mentioned matching kernels, the proposed AERK kernel can reflect multilevel alignment information through the entropic correspondence matrices. Finally, unlike the mentioned graph kernels in Section~\ref{s1.1}, the proposed AERK kernel can also simultaneously capture both global and local structural information through the quantum Shannon entropies, explaining the effectiveness of the new kernel.

\textbf{Third}, we experimentally demonstrate the graph classification performance of the proposed AERK kernel associated with the C-SVM on ten standard graph datasets. The AERV kernel significantly excels state-of-the-art graph kernel methods and graph deep learning approaches.


The remainder of this paper organizes as follows. Section \ref{s2} reviews the related works. Section \ref{s3} defines the proposed AERK kernel. Section \ref{s4} empirically evaluates the proposed QESK kernel. Section \ref{s5} gives the conclusion.

\section{Related Works}\label{s2}

In this section, we review the definitions of two classical graph kernels, that are theoretically related to this work.

\subsection{The Classical Reproducing Kernel}\label{s2.1}
We briefly review the definition of the reproducing kernel method between graphs~\cite{DBLP:journals/pr/XuJBXL18}. In mathematical theory, the Hilbert space $\mathcal{H}$ is defined as the dot product space, which is separable and complete to the norm described by the operation of dot product $\langle\cdot ,\cdot \rangle$. Moreover, $\mathcal{H}$ is also called as the Reproducing-Kernel Hilbert Space (RKHS) (i.e., a proper Hilbert space)~\cite{berlinet2011reproducing}, if it includes complex-valued functions in terms of the reproducing kernel.

\vspace{5pt}

\noindent\textbf{Definition 2.1 (The Reproducing Kernel):} The kernel function $K: \mathcal{R}\times \mathcal{R} \rightarrow \mathcal{H}$ satisfying $(a, b) \mapsto K(a, b)$ is called as the reproducing kernel in $\mathcal{H}$, iff it satisfies
\begin{itemize}
  \item{$\forall b \in \mathcal{R}$, $K(., b)\in \mathcal{H}$;}
  \item{$\forall b \in \mathcal{R}$, $\forall \phi \in \mathcal{H} $ $\langle \phi, K(., b)\rangle = \phi(b)$};
\end{itemize}
We call the second condition as the reproducing property, where the function $\phi(b)$ at point $b$ is reproduced with the dot product of $\phi$ and $K(\cdot, b)$. \hfill$\Box$

With the above scenario, Xu et al.,~\cite{DBLP:journals/pr/XuJBXL18} have developed a reproducing kernel for graphs associated with the definition of the $\mathcal{H}^1$-reproducing kernel in the $\mathcal{H}^1(\mathcal{R})$ Hilbert space. This is achieved by employing the Delta function $\sigma (b)$ that has been introduced in~\cite{DBLP:journals/ijon/XuNXAL15}. Specifically, the Delta function $\sigma (b)$ can physically represent the density of a point charge, and play a crucial role in the partial differential equation, mathematical physics, probability theory, and Fourier analysis~\cite{aronszajn1950theory}. Let $\mathcal{R}$ denote the set of real numbers, $\mathcal{Z}$ denote the set of integers, and $$\mathcal{H}^n(\mathcal{R})=\{ s(b)|s(b),s^{'}(b),s^{''}(b),\ldots,s^{(n-1)}(b)\}$$ be an absolutely continuous-function in $$\{\mathcal{R}, s^{'}(b),s^{''}(b),\ldots,s^{(n)}(b) \in L^2(\mathcal{R}) \}$$ where $n\in \mathcal{Z}^+$. The dot product in the $\mathcal{H}^n(R)$ Hilbert space is formulated as
\begin{equation}
\langle   s,t\rangle_{\mathcal{H}^n(\mathcal{R})}=\int_\mathcal{R}(\sum_{i=1}^n c_i s^{(i)} t^{(i)}) \mathrm{d} a, \forall s,t \in \mathcal{H}^n(\mathcal{R}),
\end{equation}
where $c_i$ is the coefficient of
\begin{equation}
(x+y)^n=\sum_{i=0}^n c_i x^i y^{n-i},
\end{equation}
and $i=0,1,2,\ldots,n$.

\vspace{5pt}

\noindent\textbf{Definition 2.2 (The Basic-Reproducing Kernel):} For the operator $L= 1-\frac{d^2}{d a^2}$, assume $K_1(a)$ is its basic solution, then the Basic-Reproducing Kernel (BRK) $K_1(a,b)$ in the $\mathcal{H}^1(\mathcal{R})$ Hilbert space is defined as
\begin{equation}
K_1(a,b) = K_1 (a-b) = \frac{1}{2} e^{-|a-b|},\label{reproducingK}
\end{equation}
that completely satisfies the conditions described by Definition 2.1, and is a $\mathcal{H}^1$-reproducing kernel in the $\mathcal{H}^1(\mathcal{R})$ Hilbert space. \hfill$\Box$

\vspace{5pt}

\noindent\textbf{Remarks:} In previous works~\cite{DBLP:journals/pr/XuJBXL18}, the BRK Kernel $K_1$ has been employed to further develop a novel Reproducing-based Graph Kernel (RGK) associated with the approximated von Neumann entropy~\cite{DBLP:journals/prl/HanEHW12}. For a pair of graphs $G_p$ and $G_q$, the RGK kernel $K_\mathrm{Rep}$ is defined as
\begin{align}
K_\mathrm{RGK}(G_p,G_q) &= K_1 [H_\mathrm{N}(G_p)-H_\mathrm{N}(G_q)] \nonumber \\
&= \frac{1}{2} e^{-|H_\mathrm{N}(G_p)-H_\mathrm{N}(G_q)|},
\end{align}\label{RGK}
where $H_\mathrm{N}(\cdot)$ represents the approximated von Neumann entropy of a graph structure. Since, $H_\mathrm{N}(\cdot)$ is defined based on the vertex degree matrix that is computed from the vertex adjacency matrix, it can naturally reflect global structural information of a graph. Hence, the RGK kernel is seen as a kind of global-based graph kernel that measures the kernel-based similarity between global graph structures. Moreover, since the approximated von Neumann entropy only requires time complexity $O(n^2)$ for the graph consisting of $n$ vertices, calculating the RGK kernel between a pair of graphs only requires time complexity $O(1)$. Hence, the computational complexity of the RGK kernel is $O(n^2)$, that is very fast comparing to most existing graph kernels in literatures. However, note that, the RGk kernel only focuses on the global graph characteristics, neglecting the local structural information residing in graphs. This may influence the effectiveness of the RGK kernel.


\subsection{The Classical Depth-based Matching Kernels}\label{s2.2}

\begin{figure*}
\centering
\subfigure{\includegraphics[width=1\linewidth]{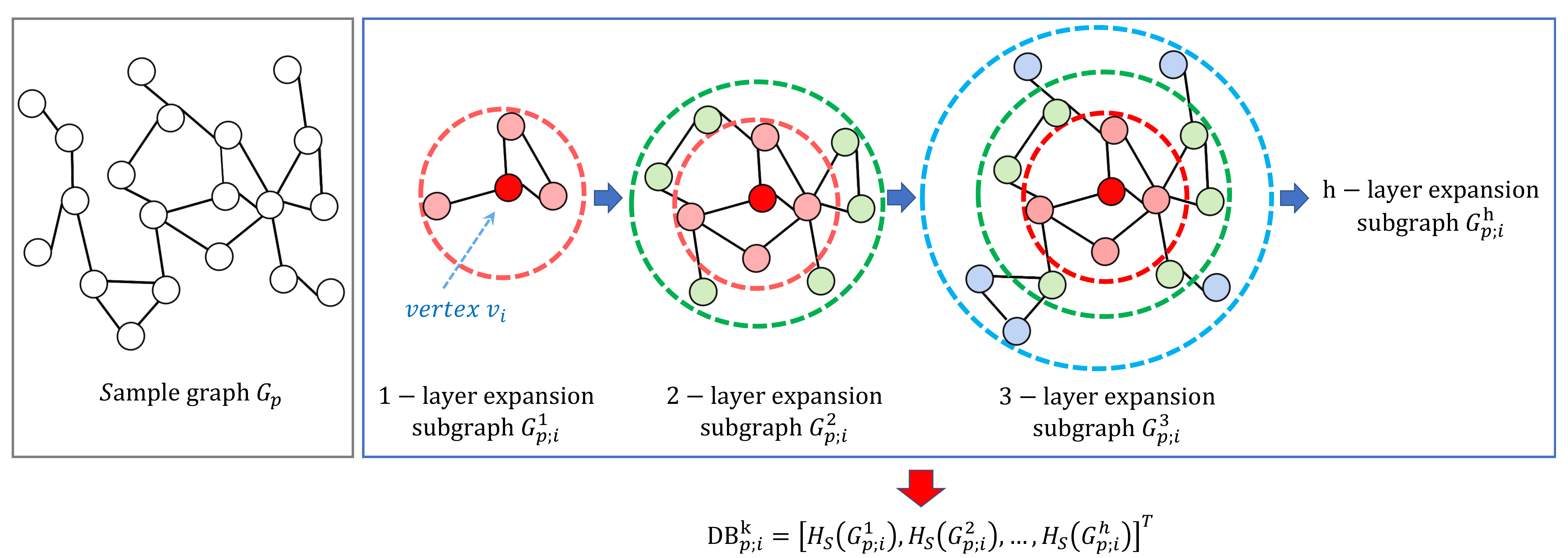}}
\vspace{-20pt}
\caption{The procedure of constructing the depth-based representations rooted at vertices. For a sample graph $G_p(V_p, E_p)$ and the associated $i$-th vertex $v_i\in V_p$ (marked with the color red), we commence by computing the $\bar{h}$-th order neighborhood set as $\mathcal{N}_{p;i}^{\bar{h}} =\{v_j \in V_p\ |\ d(v_i,v_j)\leq 1\}$, where $d(v_i,v_j)$ represents the length of the shortest path between the $j$-th vertex $v_j\in V_p$ and the $i$-th vertex $v_i\in V_p$. Then, the $\bar{h}$-layer expansion subgraph $H_S({G}_{p;i}^{\bar{h}})$ rooted at $v_i\in V_p$ can be defined as the substructure associated with the vertices in $\mathcal{N}_{i}^{\bar{h}}$ and their orogonal topological structures in the original graph $G_p$, e.g, the $1$-layer, $2$-layer and $3$-layer expansion subgraphs ${G}_{p;i}^1$, ${G}_{p;i}^2$ and ${G}_{p;i}^3$ surrounded by the red, green and blue broken line, respectively. Clearly, we can construct a family of $\bar{h}$-layer expansion subgraphs, if we vary $\bar{h}$ varies from $1$ to $h$ (i.e., $\bar{h}=1,2, \ldots, h$). Note that, if $\bar{h}$ is greater than the longest shortest path length from $v_i$ to other vertices, the $\bar{h}$-layer expansion subgraph $\mathcal{G}_{p;i}^{\bar{h}}$ is essentially the global structure of $G_p$. The resulting $h$-dimensional depth-based representation of $v_i$ is defined as ${\mathrm{DB}}^h_{p;i}=[H_S({G}_{p;i}^1),H_S({G}_{p;i}^{2}),\cdots,H_S({G}_{p;i}^{\bar{h}}),\cdots,H_S({G}_{p;i}^{h})]^T$. Here $H_S(\cdot)$ is the classical Shannon entropy of a subgraph based on classical steady-state random walks.}\label{DBR}
\vspace{-10pt}
\end{figure*}

In this subsection, we introduce the definition of a classical vertex-based matching kernel~\cite{DBLP:conf/sspr/BaiR0H14}, that computes the kernel value for the similarity between graphs by calculating the number of pairwise aligned vertices between graphs. For each pair of graphs denoted as $G_p(V_p, E_p)$ and $G_q(V_q, E_q)$, assume $\mathbf{R}^h(v_i)$ and $\mathbf{R}^h(v_j)$ are the $h$-dimensional vectorial representations of their vertices $v_i\in V_p$ and $v_j\in V_q$. We commence by computing the Euclidean distance as the affinity measure between $v_i\in V_p$ and $v_j\in V_q$, and the $(i,j)$-th entry $R(i,j)$ of the resulting affinity matrix $R\in {\mathcal{R}}^{|V_p||V_q|}$ between the graphs $G_p$ and $G_q$ can be calculated as
\begin{align}
R(i,j)=\parallel \mathbf{R}^h(v_i)-\mathbf{R}^h(v_j) \parallel.\label{AffinityM}
\end{align}
If the $(i,j)$-th entry $R(i,j)$ of $R$ is the smallest one in both $i$-th row and $j$-th column, there is a biunivocal correspondence between $v_i\in V_p$ and $v_j\in V_q$, i.e., $v_i$ and $v_j$ are aligned to each other. More specifically, we record the vertex correspondence information between $G_p$ and $G_q$ in the correspondence matrix $C\in \{0,1\}^{|V_p||V_q|}$ that satisfies
\begin{equation}
C(i,j)=\left\{
\begin{array}{cl}
1   & \mathrm{if} \  R(i,j) \ \mathrm{is \ the \ smallest \ entry \ in \ both} \\
    & i-\mathrm{th \ row \ and} \ j-\mathrm{th \ column}; \\
0   & \mathrm{otherwise}.
\end{array} \right.
\label{CoMatrix}
\end{equation}
Eq.(\ref{CoMatrix}) indicates that the vertices $v_i\in V_p$ and $v_j\in V_q$ are aligned if $C(i,j)=1$. To construct effective alignment information between $G_p$ and $G_q$, Bai et al.,~\cite{DBLP:conf/sspr/BaiR0H14} have suggested to employ depth-based representations as $h$-dimensional vectorial vertex representations. The reason of utilizing the depth-based representations is that they are defined by gauging the entropy-based complexity on a family of $\bar{h}$-layer expansion subgraphs rooted at vertices~\cite{DBLP:journals/pr/BaiH14}, where $\bar{h}$ changes from $1$ to $h$ (i.e., $\bar{h}=1,2, \ldots, h$) and the larger layer subgraph completely encapsulates the smaller layer expansion subgraph. As a result, the depth-based representation can significantly encapsulates rich entropic content flow varying from each local vertex to the global graph structure, as a function of depth. Fig.\ref{DBR} shows the computation procedure of the depth-based representation.

Finally, note that, for each pair of graphs, the vertex of a graph might be aligned to two or more vertices of the other graph. Thus, for the correspondence matrix $C$, there may be two or more entries are equal to $1$ in the same row or column. To guarantee each vertex is aligned to one vertex at most, we suggest to choose an entry with value $1$ randomly, and set other elements of that row or column of $C$ as zero. Based on the evaluation of the previous work~\cite{DBLP:conf/sspr/BaiR0H14}, this strategy will not influence the effectiveness.

\vspace{5pt}

\noindent\textbf{Definition 2.3 (The Depth-based Matching Kernel):} For the pair of graphs $G_p(V_p,E_p)$ and $G_q(V_q,E_q)$ and their associated correspondence matrix $C$ defined by Eq.\ref{CoMatrix}, the Depth-based Matching Kernel (DBMK) between $G_p$ and $G_q$ is formulated as
\begin{align}
K_\mathrm{DBMK}(G_p,G_q)=\sum_{i=1}^{|V_p|} \sum_{j=1}^{|V_q|}
C(i,j),\label{KernelMatching}
\end{align}
that counts the number of pairwise aligned vertices between $G_p$ and $G_q$. \hfill$\Box$

\vspace{5pt}

\noindent\textbf{Remarks:} Indeed, the DBMK kernel is theoretically related to a typical instance of the R-convolution graph kernels (i.e., the classical All Subgraph Kernel, ASK)~\cite{DBLP:conf/colt/GartnerFW03}, validating the effectiveness of the DBMK kernel. Specifically, for the pair of graphs $G_p$ and $G_q$, the ASK kernel is formulated as
\begin{align}
K_{\mathrm{ASK}}(G_p,G_q)=\sum_{S_p\sqsubseteq G_p} \sum_{S_q\sqsubseteq G_q}
\delta(S_p,S_q),
\end{align}
where $S_p$ and $S_q$ are any pair of subgraphs of $G_p$ and $G_q$, and
\begin{equation}
\delta(S_p,S_q)=\left\{
\begin{array}{cl}
1   & \mathrm{if} \  S_p\simeq S_q,\  \\
0   & \mathrm{otherwise}.
\end{array} \right.
\end{equation}
Here, the function $\delta$ represents a Dirac kernel. Specifically, $\delta(S_{p},S_{q})$ is equal to $1$ if $S_p$ and $S_q$ are isomorphic (i.e., $S_{p} \simeq S_{q}$), and $0$ otherwise. Since the $h$-dimensional depth-based representation ${\mathrm{DB}}^h_{p;i}$ of the vertex $v_i\in V_p$ can be theoretically considered as the vectorial representation of the associated $h$-layer expansion subgraph ${G}_{p;i}^{{h}}$ rooted at $v_i\in V_p$~\cite{DBLP:conf/sspr/BaiR0H14}. Moreover, Eq.(\ref{AffinityM}) and Eq.(\ref{CoMatrix}) indicate that ${\mathrm{DB}}^h_{p;i}$ and ${\mathrm{DB}}^h_{q;i}$ are closest to each other in the Euclidean space (i.e., ${\mathrm{DB}}^h_{p;i}$ and ${\mathrm{DB}}^h_{q;i}$ are structurally similar), if the the vertices $v_i\in V_p$ and $v_j\in V_q$ are aligned to each other. Hence, the DBMK kernel can be rewritten as
\begin{align}
K_{\mathrm{DBMK}}^h(G_p,G_q)=\sum_{S_p\sqsubseteq G_p} \sum_{S_q\sqsubseteq
G_q} \delta(S_p,S_q),\label{All_Subgraph_K}
\end{align}
where
\begin{equation}
\delta(S_p,S_q)=\left\{
\begin{array}{cl}
1   & \mathrm{if} \  S_p={G}_{p;i}^{h}\ \mathrm{and}\ S_q={G}_{q;j}^{h}, \\
    & \mathrm{and}\ v\ \mathrm{and}\ u\ \mathrm{are}\ \mathrm{aligned}, \\
0   & \mathrm{otherwise}.
\end{array} \right.\label{All_Subgraph_ISO}
\end{equation}

As a result, the DBMK kernel can be theoretically considered as an Aligned Subgraph Kernel that calculates the number of pairwise isomorphic $h$-layer expansion subgraphs around aligned vertices. In other words, the DBMK kernel integrates the local correspondence information between the isomorphic subgraphs, addressing the shortcoming of classical R-convolution graph kernels which tend to overlook structural correspondence information between substructures. Unfortunately, the DBMK kernel still suffers from some drawbacks. First, similar to the R-convolution kernels, the DBMK kernel cannot reflect global graph characteristics. Second, the DBMK kernel cannot discriminate the structural differences between different pairs of aligned vertices in terms of the global graph structures, since any pair of aligned vertices will contribute the same kernel-based similarity value (i.e., one unit value). Obviously, these drawbacks may affect the effectiveness of the DBMK kernel.

\section{The Aligned Entropic Reproducing Kernel}\label{s3}

In this section, we propose the Aligned Entropic Reproducing Kernel (AERK) for graphs. We commence by introducing the concept of the CTQW. Moreover, we employ the AMM matrix to describe the behaviour of the CTQW, and show how this AMM matrix of the CTQW allows us to define a quantum Shannon entropy for each vertex. With the quantum Shannon entropies of all vertices to hand, we define the new AERK kernel between graphs by measuring the similarity between the entropies over all pairs of aligned vertices. Finally, we discuss the theoretical advantages of the proposed AERK kernel.

\subsection{Quantum Shannon Entropies of Vertices through CTQWs}\label{s3.1}

We first introduce the concept of the CTQW~\cite{DBLP:journals/pr/EmmsWH09,rossi2015measuring}. For the CTQW evolving on a sample graph $G(V,E)$, its state space is defined over the vertex set $V$, and the corresponding basis state at each vertex $v \in V$ is defined as $\Ket{v}$ based on the Dirac notation. Here, $\Ket{.}$ is a $|V|$-dimensional orthonormal vector in a complex value Hilbert space. Since the state $\Ket{\psi(t)}$ at time $t$ is defined as a complex linear combination of these basic state vectors $\Ket{v}$, $\Ket{\psi(t)}$ can be written as $\Ket{\psi(t)} = \sum_{v\in V} \alpha_v(t) \Ket{v}$, where $\alpha_v (t) \in \mathbb{C}$ is the complex amplitude. Unlike its classical counterpart CTRW~\cite{DBLP:conf/sigmetrics/RibeiroFST11}), the evolution of the CTQW is defined based on the Schr\"{o}dinger equation, i.e., $\frac{\partial \Ket{\psi_t}}{\partial t}  = -i\mathcal{\bar{H}}\Ket{\psi_t}$. Here, $\mathcal{\bar{H}}$ denotes the system Hamiltonian accounting the entire system energy. In our approach, we propose to adopt the vertex adjacency matrix $A$ of $G$ as the required Hamiltonian.

Furthermore, the behavior of the CTQW evolving on the graph $G(V,E)$ at time $t$ is summarized with the AMM matrix~\cite{godsil2013average}, that is defined as
\begin{align}
Q_\mathrm{M}(t) &= U(t) \circ U(-t) \nonumber \\
&= e^{i\mathcal{\bar{H}}t} \circ e^{-i\mathcal{\bar{H}}t},
\end{align}
where $\circ$ represents the Schur-Hadamard product operation between $e^{i\mathcal{\bar{H}}t}$ and $e^{-i\mathcal{\bar{H}}t}$. Since $U$ is an unitary matrix, $Q_\mathrm{M}(t)$ denotes a doubly stochastic matrix and each of its entry $Q_\mathrm{M}(t)_{uv}$ corresponds to the probability of the CTQW arriving in vertex $v\in V$ at time $t$ as the CTQW departs from vertex $u\in V$. To ensure the convergence for $Q_\mathrm{M}(t)$, we compute the time-averaged AMM matrix $Q$ for the CTQW by taking the Ces\`{a}ro mean, i.e.,
\begin{equation}
Q = \lim_{T \rightarrow \infty} \int_{0}^{T} Q_M(t) dt,
\end{equation}
where each entry $Q_{vu}$ of $Q$ corresponds to the time-averaged probability of the CTQW arriving in $u\in V$ when the CTQW departs from $v\in V$. Note that, as same as $Q_\mathrm{M}(t)$, $Q$ is also a doubly stochastic matrix. Because each entry of $Q$ is a rational number~\cite{godsil2013average}, we can directly calculate $Q$ based on the Hamiltonian spectrum. Assume the adjacency matrix $A$ is the Hamiltonian $\mathcal{\bar{H}}$, $\lambda_1,\lambda_2,\ldots,\lambda_{|V|}$ be its eigenvalues, and $\mathbb{P}_j$ be its orthogonal projection on the eigenspace in association with $\lambda_j$ (i.e., $\mathcal{\bar{H}} = \sum_{j = 1}^{|V|} \lambda_j \mathbb{P}_j$), the AMM matrix $Q$ for the CTQW can be computed as
\begin{equation}
Q = \sum_{j = 1}^{|V|} \mathbb{P}_j \circ \mathbb{P}_j.
\end{equation}

\vspace{5pt}

\noindent\textbf{Definition 3.1 (The Quantum Shannon Entropy of A Vertex):} With the AMM matrix $Q$ of the graph $G(V,E)$ to hand, we define a quantum Shannon entropy $H_{\mathrm{QS}}(v)$ for each vertex $v\in V$ associated with the $v$-th row of $Q$, i.e.,
\begin{equation}
H_\mathrm{Q}(v)=-\sum_{u\in V} Q_{vu} \log Q_{vu},\label{Qentropy}
\end{equation}
where $H_\mathrm{Q}(v)$ can be seen as the structural characteristics of each local vertex $v\in V$ in terms of the CTQW departing from $v\in V$ itself. \hfill$\Box$


\vspace{5pt}

\noindent\textbf{Remarks:} The quantum Shannon entropy computed through the CTQW has some important theoretical properties. First, unlike the CTRW, the CTQW is not governed by the low Laplacian spectrum frequency, and can thus better discriminate structures of different graphs. Moreover, Bai et al., \cite{DBLP:conf/pkdd/Bai0BH14,DBLP:journals/pr/Bai0TH15} have shown that the CTQW can reflect the complicated intrinsic structural information of graph structures through the evolution of the CTQW. As a result, the quantum Shannon entropy can well characterize the graph topological information through the AMM matrix of the CTQW. Second, since each $v$-th row of the AMM matrix corresponds to an individual probability distribution of the CTQW visiting all vertices when the CTQW departs from each vertex $v$. The AMM matrix not only encapsulates the complicated global structure information through the evolution of the CTQW, but also reflects the local structural information in terms of each local vertex. Overall, the quantum Shannon entropy provides us an elegant way to develop a novel entropic graph kernel to simultaneously capture rich global and local graph structure characteristics, by measuring the similarity between the entropies of vertices.

\subsection{The Proposed AERK Kernel}\label{s3.2}

In this subsection, we give the definition for the proposed AERK kernel. To reflect multilevel correspondence information between graphs, unlike the original DBMK kernel discussed in Section~\ref{s2.2}, we propose to compute a family of different level correspondence matrices rather than a single-level correspondence matrix between graphs. Specifically, for the pair of graphs $G_p(V_p, E_p)$ and $G_q(V_q, E_q)$, assume ${\mathrm{DB}}^h_{p;i}$ and ${\mathrm{DB}}^h_{q;j}$ are the associated $h$-dimensional depth-based representations of their vertices $v_i\in V_p$ and $v_j\in V_p$ (i.e., the $h$-dimensional vectorial representations of $v_i\in V_p$ and $v_j\in V_p$), respectively. The $h$-level affinity matrix $R^h$ between the graphs $G_p$ and $G_q$ is defined as
\begin{align}
R^h(i,j)=\parallel {\mathrm{DB}}^h_{p;i} - {\mathrm{DB}}^h_{q;j} \parallel.\label{AffinityMM}
\end{align}

Different from the affinity matrix $R$ defined by Eq.(\ref{AffinityMM}), for each pair of aligned vertices $v_i\in V_p$ and $v_j\in V_q$ identified from the $h$-level affinity matrix $R^h$, we not only request the entry $R^h(i,j)$ to be the smallest one in both row $i$ and column $j$, but also request both the $h$-layer expansion subgraphs ${G}_{p;i}^{\bar{h}}$ and ${G}_{q;j}^{\bar{h}}$ for ${\mathrm{DB}}^h_{p;i}$ and ${\mathrm{DB}}^h_{q;j}$ to exist (i.e., we request $|\mathcal{N}_{p;i}^{\bar{h}}|>0$ and $|\mathcal{N}_{q;j}^{\bar{h}}|>0$). As a result, unlike the correspondence matrix $C$ defined by Eq.(\ref{CoMatrix}), the $h$-level correspondence matrix $C^h\in \{0,1\}^{|V_p||V_q|}$ recording the vertex correspondence information between $G_p$ and $G_q$ based on $R^h$ needs to satisfy
\begin{equation}
C^h(i,j)=\left\{
\begin{array}{cl}
1   & \mathrm{if} \  R^h(i,j) \ \mathrm{represents \ the \ smallest \ entry} \\
    & \mathrm{in \ both \ row} \ i \ \mathrm{and \ column} \ j, \  \mathrm{and} \\
    & |\mathcal{N}_{p;i}^{\bar{h}}|>0 \ \mathrm{and} \ |\mathcal{N}_{q;j}^{\bar{h}}|>0;\\
0   & \mathrm{otherwise}.
\end{array} \right.
\label{CoMatrixMM}
\end{equation}
When we change the parameter $h$ from $1$ to $H$ (e.g., the greatest value of $H$), we can compute a family of different $h$-level correspondence matrices as $\mathbf{C}=\{C^1, C^2, \ldots, C^h,\ldots, C^H\}$, that reflect multilevel correspondence information.

\vspace{5pt}

\noindent\textbf{Definition 3.2 (The Aligned Entropic Reproducing Kernel):} For the pair of graphs $G(V_p,E_p)$ and $G(V_q,E_q)$, and the associated $h$-level correspondence matrix set $\mathbf{C}$, the AERK kernel $\mathrm{K}_{\mathrm{AERK}}$ is proposed by computing the sum of the reproducing kernel-based similarities between the entropies over all pairs of aligned vertices, i.e.,

\begin{footnotesize}
\begin{align}
\mathrm{K}_{\mathrm{AERK}}(G_p,G_q)&=\sum_{h=1}^H \sum_{v_i\in V_p} \sum_{v_j\in V_q} C^h(i,j) \mathrm{K}_\mathrm{1}[H_{\mathrm{Q}}(v_i),H_{\mathrm{Q}}(v_j)] \nonumber \\
&=\frac{1}{2}\sum_{h=1}^H \sum_{v_i\in V_p} \sum_{v_j\in V_q} C^h(i,j)e^{-|H_{\mathrm{Q}}(v_i)-H_{\mathrm{Q}}(v_j)|},
\end{align}
\end{footnotesize}
where $\mathrm{K}_\mathrm{1}[H_{\mathrm{Q}}(v_i),H_{\mathrm{Q}}(v_j)]$ is the Basic Reproducing Kernel defined by Eq.(\ref{reproducingK}) associated with the quantum Shannon entropies $H_{\mathrm{Q}}(v_i)$ and $H_{\mathrm{Q}}(v_j)$ of the vertices $v_i\in V_p$ and $v_j\in V_q$ defined by Eq.(\ref{Qentropy}). \hfill$\Box$

\vspace{5pt}

\begin{table*}
\centering {
\footnotesize
\caption{Properties of the Proposed HAQJSK Kernels}\label{Comparison}
\vspace{-10pt}
\begin{tabular}{|c||c||c||c||c|}

  \hline
 ~Kernel Properties ~        & ~\textbf{AERK}~   &~R-convolution Kernels~ & ~Matching-based Kernels~ &  ~Global-based Kernels~\\ \hline \hline

 ~Structural Alignment~      & ~$\mathrm{\textbf{Yes}}$~  &~ $\mathrm{No}$~       & ~$\mathrm{\textbf{Yes}}$~          &  ~$\mathrm{No}$~ \\  \hline

 ~Capture Local Information~ & ~$\mathrm{\textbf{Yes}}$~  &~ $\mathrm{\textbf{Yes}}$~       & ~$\mathrm{\textbf{Yes}}$~         &  ~$\mathrm{No}$~ \\  \hline

 ~Capture Global Information~& ~$\mathrm{\textbf{Yes}}$~  &~ $\mathrm{No}$~       & ~$\mathrm{No}$~           &  ~$\mathrm{\textbf{Yes}}$~  \\  \hline

  ~Reflect Multilevel Alignments~     & ~$\mathrm{\textbf{Yes}}$~  &~ $\mathrm{No}$~        & ~$\mathrm{No}$~          &  ~$\mathrm{No}$~  \\  \hline

  ~Discriminate Different Alignments~     & ~$\mathrm{\textbf{Yes}}$~  &~ $\mathrm{No}$~        & ~$\mathrm{No}$~          &  ~$\mathrm{No}$~  \\  \hline

\end{tabular}
}
\vspace{-5pt}
\end{table*}

\subsection{Discussions of the AERK Kernel}\label{s3.3}

In this subsection, we indicate the theoretical advantages of the new developed AERK kernel by revealing the theoretical linkage to the DBMK kernel defined in Section~\ref{s2.2}. To this end, we commence by redefining the $h$-level correspondence matrix $C^h$ described by Eq.(\ref{CoMatrixMM}) as a new $h$-level entropic correspondence matrix $C^h_{\mathrm{E}}$, i.e.,

\begin{scriptsize}
\begin{equation}
C^h_{\mathrm{E}}(i,j)=\left\{
\begin{array}{cl}
\frac{1}{2}e^{-|H_{\mathrm{Q}}(v_i)-H_{\mathrm{Q}}(v_j)|}   & \mathrm{if} \  R^h(i,j) \ \mathrm{is \ the \ smallest \ entry} \\
    & \mathrm{both \ in \ row} \ i \ \mathrm{and \ in\ column} \ j,  \  \mathrm{and}  \\
    & |\mathcal{N}_{p;i}^{\bar{h}}|>0 \ \mathrm{and} \ |\mathcal{N}_{q;j}^{\bar{h}}|>0;\\
0   & \mathrm{otherwise}.
\end{array} \right.
\label{CoMatrixRED}
\end{equation}
\end{scriptsize}
For the pair of graphs $G_p(V_p,E_p)$ and $G_q(V_q,E_q)$, we say that their vertices $v_i\in V_p$ and $v_j\in V_q$ are aligned, if the entry $C^h_{\mathrm{E}}(i,j)>0$. With the matrix $C^h_{\mathrm{E}}$ to hand, the AERK kernel can be rewritten as
\begin{align}
\mathrm{K}_{\mathrm{AERK}}(G_p,G_q)&=\sum_{h=1}^H \sum_{v_i\in V_p} \sum_{v_j\in V_q} C^h_{\mathrm{E}}(i,j).\label{KernelMatchingR}
\end{align}
As a result, the definition of the proposed AERK kernel defined by Eq.(\ref{KernelMatchingR}) is similar with that of the classical DBMK kernel defined by Eq.(\ref{KernelMatching}). In other word, similar to the DBMK kernel, the AERK kernel can also be theoretically considered as a vertex-based matching kernel. The only difference between them is that the proposed AERK kernel can encapsulate multilevel correspondence information between graphs through the different $h$-level entropic correspondence matrices $C^h_{\mathrm{E}}$, where $h$ varies from $1$ to $H$ ($h\leq H$). By contrast, the classical DBMK kernel only reflects single-level correspondence information through the correspondence matrix $C$ defined by Eq.(\ref{CoMatrix}). The above theoretical relationship indicates some advantages of the proposed AERK kernel, that are shown in Table~\ref{Comparison} and briefly discussed as follows.

First, unlike the classical DBMK kernel, Eq.(\ref{CoMatrixRED}) indicates that the AERK records the correspondence information between vertices using the reproducing kernel based similarity measurement between the quantum Shannon entropies for a pair of aligned vertices. Because the visiting probability distributions of the CTQW departing from different vertices are not the same, the associated entropies of different vertices computed through the AMM matrix of the CTQW are also different. Thus, for different pairs of aligned vertices, the different entries of the $h$-level entropic correspondence matrix $C^h_{\mathrm{E}}$ will contribute different similarity values for the proposed AERK kernel, i.e., the AERK kernel can discriminate the structural differences between different pairs of aligned vertices. By contrast, for the classical DBMK kernel, $Eq.(\ref{CoMatrix})$ indicates that any pair of aligned vertices will contribute the same one unit kernel value, i.e., the classical DBMK kernel cannot identify the structural differences between different pairs of aligned vertices.

Second, like the classical DBMK kernel, the proposed AERK kernel can also be defined with the similar manners of Eq.(\ref{All_Subgraph_K}) and Eq.(\ref{All_Subgraph_ISO}). In other words, AERK kernel is also theoretically related to the ASK kernel mentioned in Section~\ref{s2.2}, and each pair of ${h}$-layer expansion subgraphs ${G}_{p;i}^{{h}}$ and ${G}_{q;j}^{{h}}$ rooted at the vertices $v_i\in V_p$ and $v_i\in V_q$ are isomorphic, if the $h$-level entropic correspondence matrix  Eq.(\ref{CoMatrixRED}) indicates that $v_i\in V_p$ and $v_i\in V_q$ are aligned to each other. As a result, the proposed AERK kernel can also be seen as an Aligned Subgraph Kernel. However, unlike the classical DBMK kernel that is based on the single-level correspondence matrix $C$ defined by Eq.(\ref{CoMatrix}) and can only identify the isomorphism between the specific $h$-layer expansion subgraphs, the proposed AERK kernel is based on the different $h$-level entropic correspondence matrices $C^h_{\mathrm{E}}$ defined by Eq.(\ref{CoMatrixRED}) and can identify the isomorphism between more different-level $h$-layer expansion subgraphs ($1\leq h\leq H$). As a result, the AERK kernel can not only resolve the problem of overlooking the correspondence information between substructures that arises in classical R-convolution graph kernels discussed in Section~\ref{s1.2}, but also reflects more structural information than the classical DBMK kernel.

Third, the proposed AERK kernel is able to address the shortcoming of only focusing on local or global graph structure information that arises in existing R-convolution graph kernels, matching-based graph kernels, and global-based graph kernels discussed in Section~\ref{s1.2}. This is due to the fact that the AERK kernel is based on measuring the similarity between the quantum Shannon entropies of aligned vertices, that can simultaneously reflect the global and local graph characteristics through the AMM matrix of the CTQW. Moreover, the proposed AERK kernel can also capture the intrinsic complicated graph structure information through the CTQW.



\subsection{Analysis of the Computational Complexity}\label{s3.4}

For a pair of graphs each consisting of $N$ vertices and $M$ edges, calculating the proposed AERK kernel between the graphs relies on four main computational steps, and these include: (1) constructing the $h$-layer depth-based representations, (2) computing the AMM matrix of the CTQW, (3) computing the $H$ different $h$-level entropic correspondence matrices, and (4) computing the kernel based on the quantum Shannon entropies of aligned vertices. Specifically, the first step depends on the shortest path computation, and thus requires time complexity $O(N^2+N* \log M)$. The second step depends on the spectrum decomposition of the adjacency matrix, and thus needs time complexity $O(N^3)$. The third step needs to compute the distance between all pairwise vertices over the graph, and thus needs time complexity $0(HN^2)$. For the pair of graphs, there are at most $N$ pairs of aligned vertices, thus the fourth step needs time complexity $O(HN)$. Thus, the entire computational complexity should be $O(N^2+N* \log M + N^3 + HN)$. Due to the fact that $N^2\gg M$ and $N\gg H$, the resulting computational complexity of the AERK kernel is $$O(N^3),$$ that indicates that the AERK kernel has a polynomial time.

\section{Experiments}\label{s4}

We make a comparison of the classification performance of the AERK kernel against both classical graph kernel methods as well as classical graph deep learning approaches on benchmark datasets.

\subsection{Benchmark Dtasets}
We evaluate the classification performance of the proposed AERK kernel on ten benchmark graph-based datasets extracted from both computer vision (CV) and bioinformatics (BIO), respectively. Specifically, the BIO datasets can be directly downloaded from~\cite{KKMMN2016}. The CV datasets are introduced in the references~\cite{DBLP:conf/dgci/BiasottiMMPSF03,DBLP:conf/cvpr/EscolanoHL11}. Table.\ref{T:GraphInformation} show the detailed statistical information of these CV and BIO datasets.
\begin{table*}
\vspace{-0pt}
\centering {
\footnotesize
\caption{Statistical Information of the Benchmark Datasets.}\label{T:GraphInformation}
\vspace{-5pt}
\begin{tabular}{|c|c|c|c|c|c|c|c|c|c|c|}

  \hline
 ~Datasets~        & ~MUTAG  ~ & ~D\&D~    & ~PTC~     & ~PPIs~    & ~CATH2~  & ~BAR31  ~ &~BSPHERE31~&~GEOD31~   & ~Shock~   & ~GatorBait~  \\ \hline \hline

 ~Max \# vertices~ & ~$28$~    & ~$5748$~  & ~$109$ ~  & ~$218$ ~  & ~$568$ ~ & ~$220$~   & ~$227$~   & ~$380$ ~  & ~$33$ ~   & ~$545$ ~ \\ \hline

 ~Min \# vertices~ &  ~$10$~   & ~$30$~    & ~$2$ ~    & ~$3$ ~    & ~$143$ ~ &  ~$41$~   & ~$43$~    & ~$29$ ~   & ~$4$ ~    & ~$239$ ~ \\ \hline

 ~Mean \# vertices~&~$17.93$~  & ~$284.4$~ & ~$25.60$~ & ~$109.63$~& ~$308.03$ ~&~$95.43$~  & ~$99.83$~ & ~$57.42$~ & ~$13.16$~ & ~$348.70$ ~\\ \hline

 ~\# graphs~       & ~$188$~   & ~$1178$~  & ~$344$ ~  & ~$219$ ~  & ~$190$ ~& ~$300$~   & ~$300$~   & ~$300$ ~  & ~$150$ ~  & ~$100$ ~\\ \hline

 ~\# classes~      & ~$2$~     & ~$2$~     & ~$2$~     & ~$5$ ~    & ~$2$ ~  & ~$15$~    & ~$15$~    & ~$15$~    & ~$10$ ~   & ~$30$ ~\\ \hline

 ~Description~     & ~$\mathrm{BIO}$~   & ~$\mathrm{BIO}$~   & ~$\mathrm{BIO}$~   & ~$\mathrm{BIO}$ ~  & ~$\mathrm{BIO}$ ~& ~$\mathrm{CV}$~    & ~$\mathrm{CV}$~    & ~$\mathrm{CV}$~    & ~$\mathrm{CV}$ ~   & ~$\mathrm{CV}$ ~\\ \hline

\end{tabular}

} \vspace{-0pt}
\end{table*}

\begin{table*}
\centering {
 \footnotesize
\caption{Graph Kernels for Comparisons.}\label{T:GKInfor}
\vspace{-5pt}
\begin{tabular}{|c|c|c|c|c|c|}

  \hline
~Kernel Methods      ~          & ~AEVK~   & ~QJSK~\cite{DBLP:journals/pr/Bai0TH15}~       & ~JTQK~\cite{DBLP:conf/pkdd/Bai0BH14}~            & ~GCCK~\cite{DBLP:journals/jmlr/ShervashidzeVPMB09}~          & ~WLSK~\cite{DBLP:journals/jmlr/ShervashidzeSLMB11} ~          \\ \hline \hline

 ~Framework~                     & ~Matching~& ~Global~    & ~R-convolution~   & ~R-convolution~ & ~R-convolution~ \\ \hline

 ~Based on CTQW~                 & ~Yes~    & ~Yes~        & ~Yes~             & ~No~            & ~No~            \\ \hline

 ~Capture Alignment Information~ & ~Yes~    & ~No~         & ~No~              & ~No~            & ~No~             \\ \hline

 ~Capture Local Information~     & ~Yes~    & ~No~         & ~Yes~             & ~Yes~           & ~Yes~            \\ \hline

 ~Capture Global Information~    & ~Yes~    & ~Yes~        & ~No~              & ~No~            & ~No~             \\ \hline

 ~Discriminate Aligned Vertices~ & ~Yes~    & ~No~         & ~No~              & ~No~            & ~No~             \\ \hline \hline

 ~Kernel Methods      ~           & ~SPGK ~\cite{DBLP:conf/icdm/BorgwardtK05}~          &~EDBMK~\cite{DBLP:journals/pr/XuBJTZL21} ~    &~ODBMK~\cite{DBLP:conf/sspr/BaiR0H14} ~   &~JSSK ~\cite{DBLP:journals/pr/BaiH16} ~         &~ISK ~\cite{DBLP:journals/pr/BaiH16} ~\\ \hline \hline

 ~Framework~                      & ~R-convolution~ &~Matching~ &~Matching~&~Local \& Global~ &~R-convolution~\\ \hline

 ~Based on CTQW~                  & ~No~            &~No~       &~No~      &~No~           &~No~\\ \hline

 ~Capture Alignment Information~  & ~No~            &~Yes~      &~Yes~     &~No~           &~No~\\ \hline

 ~Capture Local Information~      & ~Yes~           &~Yes~      &~Yes~     &~Yes~          &~Yes~\\ \hline

 ~Capture Global Information~      & ~No~            &~No~       &~No~      &~Yes~          &~No~\\ \hline

 ~Discriminate Aligned Vertices~  & ~No~            &~No~      &~No~     &~No~           &~No~\\ \hline

\end{tabular}
} \vspace{-0pt}
\end{table*}

\begin{table*}
\vspace{-0pt}
\centering {
\footnotesize
\caption{The Comparisons between Different Graph Kernel Methods.}\label{T:ClassificationGK}
\vspace{-5pt}
\begin{tabular}{|c|c|c|c|c|c|}

  \hline
 ~Datasets~  & ~MUTAG  ~        & ~D\&D~            & ~PTC~            & ~PPIs~            & ~CATH2~  \\ \hline \hline

  ~\textbf{AERK}~     & ~$\textbf{88.55}\pm.43$~  & ~$77.60\pm.47$~   & ~$59.38\pm.36$ ~ & ~$84.47\pm.56$ ~  & ~$\textbf{84.36}\pm.65$ ~ \\ \hline

  ~QJSK~     & ~$82.72\pm.44$~  & ~$77.68\pm.31$~   & ~$56.70\pm.49$ ~ & ~$65.61\pm.77$ ~  & ~$71.11\pm.88$ ~\\ \hline

  ~JTQK~     & ~$85.50\pm.55$~  & ~$\textbf{79.89}\pm.32$~   & ~$58.50\pm.39$ ~ & ~$\textbf{88.47}\pm.47$ ~  & ~$68.70\pm.69$ ~\\ \hline

  ~GCGK~     & ~$82.04\pm.39$~  & ~$74.70\pm.30$~   & ~$55.41\pm.59$ ~ & ~$46.61\pm.47$ ~  & ~$73.68\pm1.09$ ~\\ \hline

  ~WLSK~     & ~$82.88\pm.57$~  & ~$79.78\pm.36$~   & ~$58.26\pm.47$ ~ & ~$88.09\pm.41$ ~  & ~$67.36\pm.63$ ~\\ \hline

  ~SPGK~     & ~$83.38\pm.81$~  & ~$78.45\pm.26$~   & ~$55.52\pm.46$ ~ & ~$59.04\pm.44$ ~  & ~$81.89\pm.63$ ~\\ \hline

  ~EDBMK~    & ~$86.35$~        & ~$78.19$~         & ~$56.79$ ~       & ~$84.13$ ~        & ~$83.58$ ~\\ \hline

  ~ODBMK~    & ~$85.27\pm.69$~  & ~$77.85$~         & ~$55.91$ ~       & ~$83.23$ ~        & ~$82.42$ ~\\ \hline

  ~JSSK~     & ~$83.77\pm.74$~  & ~$76.32\pm.46$~   & ~$56.94\pm.43$ ~ & ~$45.04\pm.88$ ~  & ~$75.42\pm.76$ ~\\ \hline

  ~ISK~      & ~$84.66\pm.56$~  & ~$75.32\pm.35$~   & ~$\textbf{60.26}\pm.42$ ~ & ~$79.47\pm.32$ ~  & ~$67.55\pm.67$ ~\\ \hline \hline

 ~Datasets~  & ~BAR31  ~        & ~BSPHERE31~       & ~GEOD31~         & ~Shock~           & ~GatorBait~  \\ \hline \hline

  ~\textbf{AERK}~   & ~$\textbf{73.66}\pm.57$~  & ~$\textbf{62.63}\pm.41$~   & ~$\textbf{47.63}\pm.45$ ~ & ~$\textbf{46.26}\pm.74$ ~  & ~$\textbf{15.00}\pm.89$ ~ \\ \hline

  ~QJSK~     & ~$30.80\pm.61$~  & ~$24.80\pm.61$~   & ~$23.73\pm.66$ ~ & ~$40.60\pm.92$ ~  & ~$9.00\pm.89$ ~\\ \hline

  ~JTQK~     & ~$60.56\pm.35$~  & ~$46.93\pm.61$~   & ~$40.10\pm.46$ ~ & ~$37.73\pm.42$ ~  & ~$9.60\pm.87$ ~\\ \hline

  ~GCGK~     & ~$22.96\pm.65$~  & ~$17.10\pm.60$~   & ~$15.30\pm.68$ ~ & ~$26.63\pm.63$ ~  & ~$8.40\pm.83$ ~\\ \hline

  ~WLSK~     & ~$58.53\pm.53$~  & ~$42.10\pm.68$~   & ~$38.20\pm.68$ ~ & ~$36.40\pm1.00$ ~  & ~$10.10\pm.61$ ~\\ \hline

  ~SPGK~     & ~$55.73\pm.44$~  & ~$48.20\pm.76$~   & ~$38.40\pm.65$ ~ & ~$37.88\pm.93$ ~  & ~$9.00\pm.75$ ~\\ \hline

  ~EDBMK~    & ~$70.08$~        & ~$57.36$~         & ~$43.57$ ~       & ~$33.24$ ~        & ~$14.40$ ~\\ \hline

  ~ODBMK~     & ~$69.40\pm.56$~  & ~$56.43\pm.69$~   & ~$42.83\pm.50$ ~ & ~$26.73$ ~       & ~$13.76$ ~\\ \hline

  ~JSSK~     & ~$52.76\pm.47$~  & ~$43.33\pm.40$~   & ~$32.03\pm1.02$ ~ & ~$37.66\pm.80$ ~  & ~$9.20\pm.65$ ~\\ \hline

  ~ISK~      & ~$62.80\pm.47$~  & ~$52.50\pm.74$~   & ~$39.76\pm.43$ ~ & ~$39.86\pm.68$ ~  & ~$11.40\pm.52$ ~\\ \hline

\end{tabular}

} \vspace{-0pt}
\end{table*}

\subsection{Experimental Comparisons with Classical Graph Kernel Methods}
\textbf{Experimental Setups:} We compare the classification performance of the proposed AERK kernel against some classical graph kernel methods, including: (1) the Quantum Jensen-Shannon Kernel (QJSK)~\cite{DBLP:journals/pr/Bai0TH15}, (2) the Jensen-Tsallis q-difference Kernel (JTQK)~\cite{DBLP:conf/pkdd/Bai0BH14}, (3) the Graphlet-Counting Graph Kernel (GCGK)~\cite{DBLP:journals/jmlr/ShervashidzeVPMB09} associated with the graphlets of size $4$, (4) the Weisfeiler-Lehman Subtree Kernel (WLSK)~\cite{DBLP:journals/jmlr/ShervashidzeSLMB11}, (5) the Shortest Path Graph Kernel (SPGK)~\cite{DBLP:conf/icdm/BorgwardtK05}, (6) the Extended Depth-based Matching Kernel (EDBMK)~\cite{DBLP:journals/pr/XuBJTZL21} associated with the R{\'{e}}nyi entropy, (7) the Original Depth-based Matching Kernel (ODBMK)~\cite{DBLP:conf/sspr/BaiR0H14}, (8) the Jensen-Shannon Subgraph Kernel (JSSK)~\cite{DBLP:journals/pr/BaiH16}, and (9) the Entropic Isomorphic Kernel (ISK)~\cite{DBLP:journals/pr/BaiH16}. Detailed properties of these classical graph kernels for this comparison are shown in Table~\ref{T:GKInfor}. For the proposed AERK kernel, we suggest to set the parameter $H$ as $10$, this is because the computation of the required vectorial depth-based representations of vertices relies on the shortest paths. For most graphs, the $10$-layer expansion subgraphs can contain the most graph topological structures rooted at each vertex. Thus, this choice can ensure the good trade-off between the computational effectiveness and efficiency. Specifically, for each of the graph kernels, we perform the $10$-fold cross-validation method to calculate the accuracy of classification associated with the standard C-SVM~\cite{ChangLinSVM2001}. Moreover, for each kernel method on each dataset, we utilize the optimal C-SVM parameters and run the experiment for 10 times, and we calculate the averaged classification accuracies ($\pm$ standard errors) in Table~\ref{T:ClassificationGK}. It should be noted that, we follow the same parameter setup for the alternative graph kernels under comparisons, based on their original definitions in the literatures. Moreover, since some kernels have been evaluated on the datasets in the original papers, we directly use the accuracies reported in the literatures.

\textbf{Experimental Results and Analysis:} With respect to the classification performance, Table~\ref{T:ClassificationGK} indicates that the proposed AERK kernel can significantly outperform all competing graph kernels on seven of the ten datasets. Although, the AERK kernel does not perform the best on the D\&D, PTC and PPIs datasets, the AERK kernel is still competitive to most of the alternative graph kernels. The effectiveness is due to the following three reasons.

\textbf{First}, the proposed AERK kernel can encapsulate structural correspondence information between graph structures. By contrast, the alternative QJSK, JTQK, GCGK, WLSK, SPGK and ISK kernels are all the instances of R-convolution graph kernels or global-based graph kernels that tend to overlook the structural alignment information between graphs. Moreover, similar to the AERK kernel, the alternative EDBMK and ODBMK kernels are instances of vertex-based matching kernels that can integrate the structural alignment information into the kernel computation. Unfortunately, these two matching kernels focus on simply counting the number of pairwise aligned vertices, and cannot discriminate the structural difference between different pairs of aligned vertices. Moreover, the AERK kernel can reflect more multilevel correspondence information than the alternative matching kernels.  As a result, the proposed AERK kernel can compute more precise kernel-based similarities between graphs than the alternative graph kernels.

\textbf{Second}, the alternative JTQK, GCGK, WLSK, SPGK and ISK kernels are all instances of R-convolution graph kernels that compromise to adopt substructures of small sizes, due to the computational efficiency problem. Thus, these R-convolution graph kernels cannot capture global graph characteristics of global graph structures. In contrast, the alternative matching kernels EDBMK and ODBMK can only reflect local structural information through the local aligned vertices. By contrast, only the proposed AERK kernel can simultaneously capture both global and local graph characteristics, reflecting more comprehensive structural information.

\textbf{Third}, similar to the AERK kernel, the JSSK kernel can also simultaneously capture both the global and local graph structure information, by estimating the classical Jensen-Shannon Divergence between the depth-based representations rooted at the centroid vertices of graph structures. In other word, the JSSK kernel can be considered as a special R-convolution graph kernel that measures the divergence-based similarities between the entropies of each pair of $h$-layer expansion subgraphs around the centroid vertices, and thus encapsulates the structural information from the local centroid vertices to the global graphs. However, unlike the AERK kernel, the JSSK kernel cannot include the structural correspondence information between graphs. Moreover, the computation of the JSSK kernel only relies on a limited number of subgraphs around the centroid vertices, reflecting limited global and local structure information. By contrast, the AERK kernel can measure the similarities between the expansion subgraphs around any vertex, reflecting more structure information. Finally, unlike the AERK kernel, the subgraph similarity measure for the JSSK kernel relies on the Shannon entropies on the basis of the classical random walk (i.e., the CTRW). By contrast, the subgraph similarity measure for the AERK kernel relies on the quantum Shannon entropies computed from the quantum walk (i.e., the CTQW). Therefore, the proposed AERK kernel can reflect more complicated intrinsic structure information than the JSSK kernel.


\subsection{Experimental Comparisons with Classical Graph Deep Learning Approaches}

\noindent\textbf{Experimental Setups:} We further compare the classification performance of the AERK kernel to some classical graph deep learning approaches, including: (1) the EigenPooling based Graph Convolution Network (EigenPool)~\cite{DBLP:conf/icml/LeeLK19}, (2) the Specific Degree-based Graph Neural Network (DEMO-Net)~\cite{DBLP:conf/kdd/WuHX19}, (3) the Deep Graphlet Kernel (DGK)~\cite{DBLP:conf/kdd/YanardagV15}, (4) the Deep Graph Convolution Neural Network (DGCNN)~\cite{DBLP:conf/aaai/ZhangCNC18}, and (5) the Diffusion-based Convolution Neural Network (DCNN)~\cite{DBLP:conf/nips/AtwoodT16}. Note that, all these alternative graph deep learning approaches compute the classification accuracies based on the same $10$-fold cross-validation strategy with us. Thus, we straightforwardly report the accuracies from the original literatures in Table~\ref{T:ClassificationGCNN}.

\textbf{Experimental Results and Analysis:} With respect to the classification performance, Table~\ref{T:ClassificationGCNN} indicates that the proposed AERK kernel outperforms most of the deep learning approaches on two of the three datasets. Although the AERK kernel cannot achieve the best classification accuracy on the D\&D dataset, the AERK kernel is still competitive to most of the alternative deep learning approaches. In fact, the graph kernel associated with a C-SVM can be theoretically seen as a shallow learning approach, that may have lower classification performance than that of the deep learning approaches. Moreover, unlike the deep learning approach, the kernel computation cannot participate the end-to-end training process of the C-SVM. However, even under this disadvantageous context, the AERK kernel still outperforms these alternative deep learning approaches, again demonstrating the effectiveness. The reason of the effectiveness for the AERK kernel may due to the fact that, the AERK kernel can capture more complicated intrinsic graph structure characteristics through the AMM matrix of the CTQW, that can be seen as a quantum-based graph structure representation. By contrast, these alternative deep learning approaches can only reflect graph structure information through the original graph structure representation (i.e., the adjacency matrix). The experimental comparisons reveal that the framework of the alignment strategy associated with the CTQW for the proposed AERK kernel can tremendously improve the performance of graph kernel methods.

\begin{table}
\vspace{-0pt}
\centering {
\footnotesize
\caption{The Comparisons between the Proposed Graph Kernel and the Classical Graph Deep Learning Approaches.}\label{T:ClassificationGCNN}
\vspace{-5pt}
\begin{tabular}{|c|c|c|c|c}

  \hline
 ~Datasets~  & ~MUTAG  ~        & ~D\&D~             & ~PTC~     \\ \hline \hline

  ~\textbf{AERK}~     & ~$\textbf{88.55}\pm.43$~  & ~$77.60\pm.47$~    & ~$\textbf{59.38}\pm.36$ ~  \\ \hline

  ~EigenPool~& ~$79.50$~        & ~$76.60$~         & ~$-$ ~  \\ \hline

  ~DEMO-Net~ & ~$81.40$~        & ~$70.80$~         & ~$57.20$ ~   \\ \hline

  ~DGK~      & ~$82.66\pm1.45$~ & ~$\textbf{78.50}\pm0.22$~  & ~$57.32\pm1.13$~\\ \hline

  ~DCNN~     & ~$66.98$~        & ~$58.09\pm0.53$~ & ~$56.60$~   \\ \hline

  ~DGCNN~    & ~$85.83\pm1.66$~ & ~$79.37\pm0.94$~  & ~$58.59\pm2.47$~ \\ \hline

\end{tabular}
} \vspace{-0pt}
\end{table}

\section{Conclusions}\label{s5}

In this work, we have developed a novel AERK kernel associated with the CTQW for graph classification. To this end, we have employed the AMM matrix of the CTQW to compute the quantum Shannon entropies of vertices, and the resulting AERK kernel can be defined by computing the sum of the reproducing kernel-based similarities between the entropies of all pairs of aligned vertices. The proposed kernel cannot only encapsulate structural alignment information between graphs, but also discriminate the structural difference between aligned vertices. Moreover, the proposed AERK kernel can simultaneously capture both the global and local graph structure characteristics. As a result, the AERK kernel can significantly address the theoretical drawbacks of state-of-the-art R-convolution graph kernels, vertex-based matching kernels and global-based graph kernels. Experimental results validate the effectiveness of the AERK kernel.




\balance


\bibliographystyle{IEEEtran}
\bibliography{example_paper}

\end{document}